\documentclass{article}

\newcommand{\cT}{\mathcal{T}}
\newcommand{\cM}{\mathcal{M}}
\newcommand{\cD}{\mathcal{D}}
\newcommand{\set}[1]{\left\{#1\right\}}
\newcommand{\abs}[1]{\left|#1\right|}
\newcommand{\R}{\mathbb{R}}

\newcommand{\ybar}{\bar{y}}
\newcommand{\yhat}{\hat{y}}

\newcommand{\precision}[1]{\mathrm{Precision}(#1)}
\newcommand{\recall}[1]{\mathrm{Recall}(#1)}
\newcommand{\E}{\mathbb{E}}
\newcommand{\dist}[1]{\mathrm{dist}\left(#1\right)}
\newcommand{\defeq}{\stackrel{\Delta}{=}}
\newcommand{\elltilde}{\tilde{\ell}}
\newcommand{\algname}{CHAMP\xspace}
\newcommand{\algnameasymm}{CHAMP-Asymmetric\xspace}
\newcommand{\algnamehard}{CHAMP-H\xspace}
\newcommand{\algnamesoft}{CHAMP-S\xspace}

\newcommand{\distmax}{\mathrm{dist}_{\textrm{max}}}
\newcommand{\red}[1]{{\color{red} #1}}
\newcommand{\blue}[1]{{\color{blue} #1}}
\newcommand{\hard}{\textrm{hard}}
\newcommand{\soft}{\textrm{soft}}
\newcommand{\distharmonic}[1]{\textrm{dist}_\textrm{harmonic}(#1)}

\usepackage{lipsum}
\newif \iflong

\usepackage[preprint, nonatbib]{neurips_2022}

\usepackage[utf8]{inputenc} 
\usepackage[T1]{fontenc}    
\usepackage{hyperref}       
\usepackage{url}            
\usepackage{booktabs}       
\usepackage{amsfonts,amsmath,amssymb}       
\usepackage{nicefrac}       
\usepackage{microtype}      
\usepackage{xcolor}         
\usepackage{todonotes}
\usepackage{multirow}
\usepackage{enumitem}
\usepackage{caption} 
\usepackage{listings}
\usepackage{color}

\definecolor{dkgreen}{rgb}{0,0.6,0}
\definecolor{gray}{rgb}{0.5,0.5,0.5}
\definecolor{mauve}{rgb}{0.58,0,0.82}

\usepackage{bm}
\usepackage{subcaption}
\usepackage{xspace}
\newcommand{\champ}{CHAMP \xspace}
\title{All mistakes are not equal: Comprehensive Hierarchy Aware Multi-label Predictions (CHAMP)
}

\author{
  Ashwin Vaswani \\
  Google Research India\\
  \texttt{ashwinvaswani@google.com} \\
  \And
  Gaurav Aggarwal \\
  Google Research India\\
  \texttt{gauravaggarwal@google.com} \\
  \And
  Praneeth Netrapalli \\
  Google Research India\\
  \texttt{pnetrapalli@google.com} \\
  \And
  Narayan G Hegde \\
  Google Research India\\
  \texttt{hegde@google.com} \\
}

\begin{document}

\maketitle

\begin{abstract}
This paper considers the problem of Hierarchical Multi-Label Classification (HMC), where (i) several labels can be present for each example, and 
(ii) labels are related via a domain-specific hierarchy tree.
Guided by the intuition that all mistakes are not equal, we present Comprehensive Hierarchy Aware Multi-label Predictions (CHAMP), a framework that penalizes a misprediction depending on its severity as per the hierarchy tree. While there have been works that apply such an idea to single-label classification, to the best of our knowledge, there are limited such works for multilabel classification focusing on the severity of mistakes. The key reason is that there is no clear way of quantifying the severity of a misprediction a priori in the multilabel setting. In this work, we propose a simple but effective metric to quantify the severity of a mistake in HMC, naturally leading to CHAMP. Extensive experiments on six public HMC datasets across modalities (image, audio, and text) demonstrate that incorporating hierarchical information leads to substantial gains as CHAMP improves both AUPRC (2.6\% median percentage improvement) and hierarchical metrics (2.85\% median percentage improvement), over stand-alone hierarchical or multilabel classification methods. Compared to standard multilabel baselines, CHAMP provides improved AUPRC in both robustness (8.87\% mean percentage improvement) and less data regimes. Further, our method provides a framework to enhance existing multilabel classification algorithms with better mistakes (18.1\% mean percentage improvement).

\end{abstract}

\section{Introduction}\label{sec:intro}

Many real-world prediction tasks have implicit and explicit relationships among labels that can be encoded by co-occurrence and graphical structures. Multiple labels per data point (co-occurrence) and hierarchical relationships (graphical) cover a wide spectrum of possible ways to express complex relations (is-a, part-of) among labels which can provide domain-specific semantics. For example, popular data streams like movies (genre), blogs (topics), natural images (objects), biopsy images (tumor grades), and others provide multilabel hierarchical information. Our own paper which talks about hierarchy, multilabel, classification topics is an example in topic classification task. Even single label datasets like Imagenet have shown better classification performance when trained with hierarchy \cite{https://doi.org/10.48550/arxiv.1811.07125} and multilabel \cite{https://doi.org/10.48550/arxiv.2101.05022} objectives individually. 

Intuitively, Hierarchical Multilabel Classification (HMC) offers advantages by giving strictly more domain-specific information over purely multilabel or hierarchical single label settings. HMC could also improve performance in detection, ranking, and generative modeling tasks.

Literature is rich in both hierarchy and multilabel classification individually. Single label hierarchical (SLH) algorithms encode domain knowledge using tree structures with labels to improve classification performance and make better errors by penalizing a misprediction depending on its severity as per the given hierarchy tree. To achieve this, popular ideas are hierarchy aware custom loss functions \cite{https://doi.org/10.48550/arxiv.1912.09393}, label embeddings \cite{giunchiglia2020coherent} and custom model architectures \cite{app11083722}. Extending these ideas from single-label hierarchical systems to multilabel scenarios is difficult primarily because these ideas are tailor-made for multi-class classification loss functions encountered in SLH but not for multiple separate classification problems as encountered in HMC.

Consequently, HMC systems designed in both pre and post-deep learning eras have been domain-specific \cite{pmlr-v8-cesa-bianchi10a,10.5555/3104482.3104485}. We observe no prior works on audio or vision HMC datasets to the best of our knowledge. HMC works are further limited by assumptions on tree structures / level-wise semantics \cite{6121678}, custom model architectures \cite{wehrmann2018hierarchical} and a lack of HMC metrics that consider the severity of errors. Earlier works that use both hierarchy and multilabel information focus on improving the overall classification performance without focusing on making better mistakes. The key reason is that there is no a priori clear way of quantifying the severity of a misprediction in the multilabel setting. One of our key contributions is to use the notion of a sphere of influence that can help quantify the severity of a misprediction.
At a high level, given a set of ground truth labels, the sphere of influence of each of these ground truth labels captures the set of all labels closest to that ground truth label.
With this notion, we rate the severity of a misprediction depending on the distance of the predicted label to the corresponding ground truth label to whose sphere of influence the predicted label belongs. To the best of our knowledge, ours is the first work for multilabel classification that proposes a notion of severity of a misprediction. The proposed method can work with global, local, hybrid and other previous HMC algorithms \cite{6121678}, \cite{5360345}, \cite{10.1145/3178876.3186005}, \cite{pmlr-v80-wehrmann18a}. Our work is domain agnostic and provides freedom to capture a wide range of label relations. For example,  
there is no need to explicitly define co-occurrence among labels,
no restriction on semantics associated with nodes at the same level in the hierarchy, and ground truth nodes can be leaf and non-leaf nodes.
This generalisability provides wider applicability of our work across a range of HMC domains and datasets.

The key contributions of this work are as follows:
\begin{itemize}[noitemsep, topsep=0.5pt]
    \item We propose a simple but effective metric to quantify the severity of a mistake in HMC, naturally leading to \champ.
    \item We showcase improved performance (AUPRC) and make better mistakes consistently using \champ on six well-known HMC datasets across text, image, and audio.
    \item We motivate hierarchical and multilabel modeling by showing the advantages of learning with lesser data, robustness to noise, and resilience to skewed label distributions.
    \item Further, our method provides a framework to supplement existing and future hierarchical and multilabel classification algorithms encouraging them to make better mistakes.
\end{itemize}

The rest of this paper is organized as follows. In Section 2, we introduce the notation and terminology. In Section 3, we present our approach. Experimental results are presented in Section 4, followed by discussion in Section 5, while the related work is discussed in Section 6. The last section gives some concluding remarks.

\section{Preliminaries and Problem Setting}
\label{sec:terminology}
We are given a set of labeled training examples $\set{(x_i,y_i): i=1,\cdots,n}$, where $x_i \in \R^d$ is the input example and $y_i \in \set{0,1}^L$ is the associated label vector, each $(x_i,y_i)$ drawn from an underlying distribution $\cD$ on $\R^d \times \set{0,1}^L$. Here $L$ denotes the total number of labels/classes. We use the term label/class interchangeably. In addition to the training dataset, we are given a hierarchy tree $\cT$ with $L$ nodes where each node corresponds to one of the classes. Our goal is to train a prediction model $\cM$ that takes $x$ as input and outputs an $L$-dimensional real valued score vector $\ybar \in [0,1]^L$. This real valued score vector $\ybar$ is converted to a Boolean prediction vector $\yhat \in \set{0,1}^L$ using a scalar threshold $\tau$. So, given a scalar threshold $\tau \in [0,1]$, the final prediction for class $j \in [L]$ is $\yhat_j = 1$ if $\ybar_j \geq \tau$ and $\yhat_j=0$ otherwise.

\subsection{Metrics}
We now present the metrics we use to evaluate our method and existing methods.
Given a model $\cM$ and threshold $\tau$, the precision and recall of class $j$ is given by:
\begin{align*}
    \precision{j}= \E_{(x,y)\sim \cD}\left[y_j=1 \middle \vert \yhat_j = 1\right] \quad \mbox{ and }    \quad  \recall{j}= \E_{(x,y)\sim \cD}\left[\yhat_j=1 \middle \vert y_j = 1\right].
\end{align*}
The overall precision and recall are then given by taking an average over all the $L$ labels.

Our first key metric is \textbf{Area under the precision-recall curve (AUPRC)}, which, as the name suggests, is given by the Area under the curve traced by precision vs. recall, as the threshold $\tau$ changes. \textbf{Precision@k} is given by the average precision of the top $k$ predictions on each example, ranked according to their scores. \textbf{F1@k} is given by the harmonic mean of the average precision and average recall over the top $k$ predictions on each example ranked according to their scores.

Many other multilabel metrics like hamming loss \cite{article1} are used, which are similar and not discussed here—the metrics widely used in multilabel classification do not consider the hierarchy over labels.

We now introduce the notion of a sphere of influence based on the hierarchy tree, which will help us quantify the severity of a misprediction.
Given two labels $j,j' \in [L]$, we use $\dist{j,j'}$ to denote the distance between the corresponding labels on the hierarchy tree $\cT$. Given a subset of labels $S \subseteq [L]$, the distance of a label $j$ to $S$ is given by $\dist{j,S}\defeq \min_{j'\in S} \dist{j,j'}$. The sphere of
influence of a label $j'\in S$ is the set $\set{j \in [L]: \dist{j,j'} = \dist{j,S}}$, or in other words those labels $j\in [L]$ such that among all $k\in S$, $j'$ minimizes the distance to $j$. Given a set $S$, we denote the sphere of influence of label $j'\in S$ by $I_{S}(j')$. This concept is illustrated in Figure~\ref{fig:tree_diagram}.

\textbf{Average Hierarchical Cost (AHC)}: Given a prediction $\yhat \in \set{0,1}^L$, this is given by $\sum_j \yhat_j \cdot \dist{j,S} / |S|$, where $S=\set{j': y_{j'}=1}$ is the set of ground truth labels. Note that this can also be equivalently written as $\sum_j \yhat_j \dist{j,k_j}/|S|$ where $j \in I_S(k_j)$. \cite{garnot2021leveraging}

\textbf{Depth of Lowest Common Ancestor (LCA)}:
Given a predicted label $j$, and the set of ground truth labels $S$, we first compute $j'$ such that 
$j\in I_S(j')$,
and compute the depth, computed from the root, of the least common ancestor for $j$ and $j'$ on the hierarchy tree~\cite{10.1007/978-3-642-15555-0_6,5206848}.
As remarked in \cite{5206848}, this measure should be thought of in logarithmic
terms, as the number of confounded classes is exponential in the height of the ancestor. 

\textbf{Hierarchical Precision (HP)}: Given a set of predicted labels as well as a set of ground truth labels for any given example, HP computes precision after expanding both the sets to include all the labels on the path to the root. More concretely, if we use $\hat{P}_i$ and $\hat{T}_i$ to be the set of predicted and ground truth labels of the $i^{\textrm{th}}$ example respectively (including all the nodes on the path to the root), then hierarchical precision is given by $HP \defeq \frac{\sum_i \abs{\hat{P}\cap \hat{T}}}{\sum_i \abs{\hat{P}}}$.

\cite{article}, \cite{RAMIREZCORONA2016179} and \cite{article1} provide a more comprehensive description of some of the above metrics.

\section{Method}
\label{sec:methods}
A simple approach to multi-label classification is to solve $L$ binary classification problems, one for each label using the binary cross entropy (BCE) loss given by:
\begin{align}
    \ell_{(x,y)}(\cM) = - \sum_{j=1}^L \left\{y_j \log \yhat_j + (1-y_j) \log \left(1-\yhat_j\right) \right\},\label{eqn:bce}
\end{align}
where $\yhat$ is the score output by model $\cM$ on input $x$. Our approach is motivated by a simple but important intuition: mispredictions must not be penalized uniformly but instead depending on their severity. While all false negatives are equally severe since they are ground truth labels, the severity of a false positive depends on the closeness of the predicted false positive with ground truth labels. So, we propose the following modified BCE loss function:
\begin{align}
    \elltilde_{(x,y)}(\cM) = - \sum_{j=1}^L \left\{y_j \log \yhat_j + \left(1+s_S(j)\right)(1-y_j) \log \left(1-\yhat_j\right) \right\},\label{eqn:mod-bce}
\end{align}
where $S$ denotes the ground truth labels i.e., $S=\set{j' : y_{j'}=1}$ and $s_S(j)$ denotes the severity of a false positive prediction on label $j$.
Since we have access to the hierarchy tree on the labels, we estimate the severity coefficients $s_S(j)$ using the distance of $j$ to the ground truth labels $S$ on the hierarchy tree $\cT$. We consider two versions of the severity metric: a hard version and a soft version: 
\begin{align}
    s_S^{\hard}(j) \defeq \beta \cdot \frac{\dist{j,S}}{\distmax} \mbox{and} \label{eqn:loss-hard} \\
    s_S^{\soft}(j) = \beta \cdot \frac{\distharmonic{j,S}}{\distmax}. \label{eqn:loss-soft}
\end{align}
Here, $\distmax \defeq \max_{k,k'\in [L]} \dist{k,k'}$ denotes the maximum distance between any two labels on the hierarchy tree, $\distharmonic{j,S} \defeq \frac{\abs{S}}{\sum_{j'\in S} \dist{j,j'}^{-1}}$ denotes the Harmonic mean of distances from $j$ to each ground truth label in $S$ and $\beta$ is a scaling parameter (a hyperparameter).
\longtrue
\iflong 
While the hard version considers only distance to the closest ground-truth label, the soft version considers distances to all ground truth labels. It is consequently more robust to outliers in the ground truth labels. This naturally leads to two versions of our \algname algorithm: \algnamehard which uses $s_S^{\hard}$~\eqref{eqn:loss-hard} in the loss function~\eqref{eqn:mod-bce} and \algnamesoft which uses $s_S^{\soft}$~\eqref{eqn:loss-soft}.
\begin{figure*}[ht]
	\centering
	\includegraphics[width=1\textwidth, height=60mm]{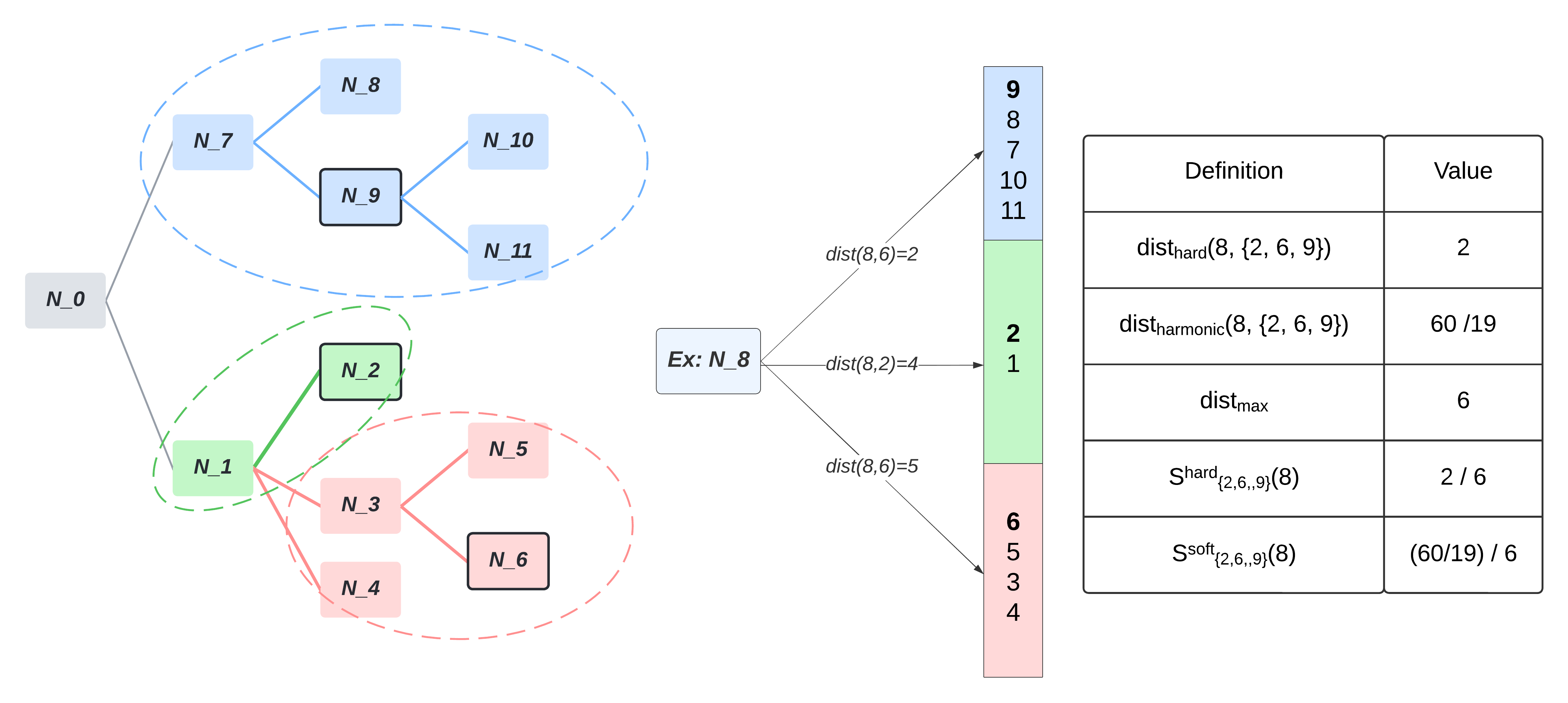}
	\caption{An example illustrating the severity metrics~\eqref{eqn:loss-hard} and~\eqref{eqn:loss-soft} used in \algnamehard and \algnamesoft respectively. The ground truth labels are $S= \set{N_2, N_6, N_9}$. The sphere of influence of each of the ground truth labels is indicated by the color as well as using an ellipse. The label under consideration is $j=N_8$. The distances of $N_8$ to $N_2, N_6$ and $N_9$ are $4, 5$ and $2$ respectively. Different versions of the distance $\dist{j,S}$ and $\distharmonic{j,S}$, maximum distance $\distmax$ as well as $s_S^{\hard}$ and $s_S^{\soft}$ are computed. We set the scaling parameter $\beta=1$ for illustration.}
	\label{fig:tree_diagram}
\end{figure*}

\section{Results}
\label{sec:results}
In this section, we present our main results evaluating \algname on six different datasets spanning across vision, audio and text. Our results demonstrate that:
\begin{itemize}[noitemsep, topsep=0.5pt]
    \item \algname provides substantial improvements over the standard BCE loss for multilabel classification, where hierarchy is not used as well as a \emph{multiclass} version of \algname where multiple labels are replaced by single labels.
    \item \algname provides improved robustness to a large set of popular corruptions for images.
    \item \algname reduces sample complexity.
    \item Finally, the notion of severity of a misprediction used in \algname can be used along with other multilabel loss functions.
\end{itemize}
\subsection{Datasets}
\begin{table}
\resizebox{\linewidth}{!}{
\begin{tabular}{lllllllllll}
\toprule
\textbf{Type}          & \textbf{Dataset}                                        & \textbf{Train} & \textbf{Test} & \textbf{\begin{tabular}[c]{@{}l@{}}\bm{$\mu_s$} \end{tabular}} & \textbf{\begin{tabular}[c]{@{}l@{}}\bm{$\mu_l$}\end{tabular}} & \textbf{\begin{tabular}[c]{@{}l@{}}\bm{$\max_l$}\end{tabular}} & \textbf{BF} & \textbf{D} & \textbf{Leaf} & \textbf{\begin{tabular}[c]{@{}l@{}}NL\end{tabular}} \\ \midrule
\multirow{3}{*}{Image} & Food201                                                 & 35242          & 15132         & 334                                                                                                                                     & 1.91                                                           & 9                                                             & 6.4         & 4          & 201           & 35                                                           \\ 
                       & \begin{tabular}[c]{@{}l@{}}COCO\end{tabular}      & 1,18,287       & 40,670        & 4287                                                                                                                                  & 2.89                                                           & 18                                                            & 6.57        & 3          & 80            & 14                                                           \\ 
                       & \begin{tabular}[c]{@{}l@{}}OpenImages V4\end{tabular} & 17,43,042      & 1,25,436      & 7384                                                                                                                              & 2.53                                                           & 19                                                            & 6.81        & 5          & 526           & 73                                                           \\ \midrule
\multirow{2}{*}{Text}  & RCV1                                                    & 23,149         & 7,81,265      & 729                                                                                                                                 & 3.24                                                           & 17                                                            & 4.71        & 4          & 82            & 21                                                           \\  
                       & NYT                                                     & 12,79,092      & 5,47,863      & 69,314                                                                                                                           & 2.52                                                           & 14                                                            & 4.17        & 4          & 91            & 27                                                           \\ \midrule
Audio                  & \begin{tabular}[c]{@{}l@{}}FSDK Audio\end{tabular}   & 4970           & 4480          & 72                                                                                                                                 & 1.4                                                            & 6                                                             & 2.53        & 5          & 80            & 49                                                           \\ 
\bottomrule
\end{tabular}
}
\caption{Detailed information about the six public datasets used in our experiments. Train and Test correspond to the number of train and test samples in the datasets, respectively. $\mu_s$ corresponds to the mean number of samples per label. $\mu_l$ and $\max_l$ correspond to the mean number of labels per sample and the maximum number of labels per sample. Finally, BF, D, Leaf, and NL represent the branching factor, depth, and the number of leaf and non-leaf nodes in the tree.}
\label{tab:dataset_info}
\end{table}
We evaluate our approach on six different public HMC  datasets across image, text, and audio modalities. We perform multilabel image classification on OpenImages V4 \cite{Kuznetsova_2020}, Food 201 \cite{7410503}, and MS-COCO 2017 \cite{https://doi.org/10.48550/arxiv.1405.0312} datasets, multilabel text classification on Reuters Corpus Volume 1 (RCV1) \cite{10.5555/1005332.1005345} which is a newswire dataset of the articles collected between 1996-1997 from Reuters and New York Times articles (NYT) \cite{Sandhaus2008Nyt} which contains articles from New York Times published between January 1st, 1987 and June 19th, 2007 along with multilabel audio classification on FSDKaggle2019 (Freesound Audio Tagging 2019) dataset \cite{https://doi.org/10.48550/arxiv.1906.02975}. The hierarchies for OpenImages, COCO, RCV1, NYT, and FSDK are provided by the authors themselves, whereas the hierarchy for Food201 was adapted based on the hierarchy given in \cite{10.1145/2964284.2967205}. Detailed label and hierarchical information of each dataset can be found in Table~\ref{tab:dataset_info}. If a given node in the tree is marked as ground truth, we also consider all of its parents present in the sample for each dataset. These six datasets cover a diverse setting with respect to data distribution and type of hierarchy, helping us conduct a comprehensive analysis. 

\subsection{Baselines}
We primarily compare \algname with a baseline model for each dataset consisting of $L$ different classifiers on top of a well-known feature extractor backbone. Each classifier is a logistic classifier trained with BCE loss~\eqref{eqn:bce}. In order to make the comparison fair, we also give different weights to positives and negatives in the BCE loss and perform hyperparameter tuning over the relative weight of positives to negatives.
The reason we limit our baselines to (variants of) standard BCE loss is that (i) existing multilabel hierarchical algorithms are domain and architecture-specific and are not general purpose~\cite{wehrmann2018hierarchical}, (ii) converting multilabel classification to multiclass classification and then using hierarchical multiclass formulations leads to a large drop in accuracies as we show in Section~\ref{sec:results-singlelabel}, and (iii) \algname can be easily augmented with more sophisticated multilabel loss functions and leads to further improvements over those methods as we show in Section~\ref{sec:results-augmentotherloss}.

\subsection{Training configuration}
\textbf{Image classification: }We conduct experiments across two different backbones (Efficientnetv2S \cite{https://doi.org/10.48550/arxiv.2104.00298} and Mobilenetv2 \cite{https://doi.org/10.48550/arxiv.1801.04381}) initialized using imagenet pre-trained weights followed by a dropout layer with probability 0.4 and a final sigmoid classification layer. We use an image size of 224 x 224 and standard augmentations like Horizontal Flip, Rotation, Contrast, Translation, and Zoom. It is important to note that state-of-the-art approaches \cite{https://doi.org/10.48550/arxiv.2009.14119} use larger image sizes for training (448 x 448), autoaugment \cite{https://doi.org/10.48550/arxiv.1805.09501} and cutout \cite{https://doi.org/10.48550/arxiv.1708.04552} for augmentations, one-cycle learning rate scheduler \cite{https://doi.org/10.48550/arxiv.1708.07120} among other tricks. We are not trying to compete with the state-of-the-art results on multi-label classification and want to demonstrate the value of adding hierarchical knowledge. 

\textbf{Text classification:} We conduct experiments by using small bert \cite{https://doi.org/10.48550/arxiv.1908.08962} (uncased, L=2, H=768, A=12) and \cite{https://doi.org/10.48550/arxiv.2004.02984} embeddings as base extractor initialised by weights trained on Wikipedia \cite{10.1145/2629489} and Bookscorpus datasets \cite{Zhu_2015_ICCV}. All text sentences are first converted to lower case.

\textbf{Audio classification:} We convert the audio wav files into Mel spectrogram with sampling rate of 44100, number of Mel bands as 347,
            length of the FFT window as 2560, lowest frequency as 20, and highest frequency as 44100//2.

All models: image, text, and audio are trained for 25 epochs using Adam \cite{https://doi.org/10.48550/arxiv.1412.6980} optimizer, an initial learning rate of 1e-4, decay as 4e-5, and a batch size of 32. While training, we reduce the learning rate on plateau with a factor of 0.2 and patience of 5 epochs. We tuned the hyperparameter $\beta$ (between 0 and 1) using validation auprc performance. Data splits are provided in the supplementary material. All experiments are conducted thrice on V100 GPUs, and mean scores are reported in the tables and figures. We have found most standard deviation scores to be <0.005.

\subsection{Main results}

\begin{table}[ht]
\centering
\resizebox{\linewidth}{!}{
\begin{tabular}{llllllll}
\toprule
\textbf{Dataset}                                                                     & \textbf{Experiment}                                           & \textbf{AUPRC}  \red{$\uparrow$}                                                            & \textbf{P}@5 \red{$\uparrow$}                                                               & \textbf{F1}@5 \red{$\uparrow$}                                                              & \textbf{LCA} \red{$\uparrow$}                                                               & \textbf{AHC}    \blue{$\downarrow$}                                                            & \textbf{HP}    \red{$\uparrow$}                                                             \\ \midrule
\multirow{3}{*}{\begin{tabular}[c]{@{}l@{}}OpenImages v4\end{tabular}} & BCE                                                  & \begin{tabular}[c]{@{}l@{}}0.342 +-  0.009\end{tabular}          & \begin{tabular}[c]{@{}l@{}}0.217 +-  0.004\end{tabular}          & \begin{tabular}[c]{@{}l@{}}0.288 +-  0.008\end{tabular}          & \begin{tabular}[c]{@{}l@{}}1.556 +-  0.012\end{tabular}          & \begin{tabular}[c]{@{}l@{}}0.878 +-  0.023\end{tabular}          & \begin{tabular}[c]{@{}l@{}}0.603 +- 0.005\end{tabular}           \\ 
                                                                            & \begin{tabular}[c]{@{}l@{}}\algname -H (0.9)\end{tabular} & \textbf{\begin{tabular}[c]{@{}l@{}}0.564 +-  0.002\end{tabular}} & \begin{tabular}[c]{@{}l@{}}0.280 +-  0.001\end{tabular}          & \begin{tabular}[c]{@{}l@{}}0.373 +-  0.001\end{tabular}          & \begin{tabular}[c]{@{}l@{}}1.852 +-  0.006\end{tabular}          & \begin{tabular}[c]{@{}l@{}}0.472 +-  0.002\end{tabular}          & \begin{tabular}[c]{@{}l@{}}0.714 +-  0.001\end{tabular}          \\ 
                                                                            & \begin{tabular}[c]{@{}l@{}}\algname -S (0.5)\end{tabular} & \begin{tabular}[c]{@{}l@{}}0.559 +-  0.006\end{tabular}          & \textbf{\begin{tabular}[c]{@{}l@{}}0.283 +-  0.001\end{tabular}} & \textbf{\begin{tabular}[c]{@{}l@{}}0.375 +-  0.002\end{tabular}} & \textbf{\begin{tabular}[c]{@{}l@{}}1.920 +-  0.014\end{tabular}} & \textbf{\begin{tabular}[c]{@{}l@{}}0.454 +-  0.021\end{tabular}} & \textbf{\begin{tabular}[c]{@{}l@{}}0.726 +-  0.007\end{tabular}} \\ \midrule
\multirow{3}{*}{\begin{tabular}[c]{@{}l@{}}Food201\end{tabular}}         & BCE                                           & \begin{tabular}[c]{@{}l@{}}0.577 +-  0.001\end{tabular}          & \begin{tabular}[c]{@{}l@{}}0.485 +-  0.002\end{tabular}          & \begin{tabular}[c]{@{}l@{}}0.462 +-  0.001\end{tabular}          & \begin{tabular}[c]{@{}l@{}}1.415 +-  0.013\end{tabular}          & \begin{tabular}[c]{@{}l@{}}0.577 +-  0.017\end{tabular}          & \begin{tabular}[c]{@{}l@{}}0.743 +-  0.006\end{tabular}          \\ 
                                                                            & \begin{tabular}[c]{@{}l@{}}\algname -H (0.9)\end{tabular} & \begin{tabular}[c]{@{}l@{}}0.585 +-  0.001\end{tabular}          & \begin{tabular}[c]{@{}l@{}}0.522 +-  0.003\end{tabular}          & \textbf{\begin{tabular}[c]{@{}l@{}}0.486 +-  0.001\end{tabular}} & \begin{tabular}[c]{@{}l@{}}1.486 +-  0.008\end{tabular}          & \begin{tabular}[c]{@{}l@{}}0.462 +-  0.014\end{tabular}          & \begin{tabular}[c]{@{}l@{}}0.788 +-  0.006\end{tabular}          \\
                                                                            & \begin{tabular}[c]{@{}l@{}}\algname -S (0.3)\end{tabular} & \textbf{\begin{tabular}[c]{@{}l@{}}0.593 +-  0.003\end{tabular}} & \textbf{\begin{tabular}[c]{@{}l@{}}0.528 +-  0.006\end{tabular}} & \textbf{\begin{tabular}[c]{@{}l@{}}0.486 +-  0.002\end{tabular}} & \textbf{\begin{tabular}[c]{@{}l@{}}1.557 +-  0.021\end{tabular}} & \textbf{\begin{tabular}[c]{@{}l@{}}0.316 +-  0.038\end{tabular}} & \textbf{\begin{tabular}[c]{@{}l@{}}0.861 +-  0.016\end{tabular}} \\ \midrule
\multirow{3}{*}{COCO}                                                       & BCE                                           & \begin{tabular}[c]{@{}l@{}}0.780 +-  0.001\end{tabular}          & \begin{tabular}[c]{@{}l@{}}0.762 +-  0.008\end{tabular}          & \begin{tabular}[c]{@{}l@{}}0.566 +-  0.002\end{tabular}          & \begin{tabular}[c]{@{}l@{}}1.994 +-  0.006\end{tabular}          & \begin{tabular}[c]{@{}l@{}}0.266 +-  0.008\end{tabular}          & \begin{tabular}[c]{@{}l@{}}0.820 +-  0.005\end{tabular}          \\ 
                                                                            & \begin{tabular}[c]{@{}l@{}}\algname -H (0.1)\end{tabular} & \begin{tabular}[c]{@{}l@{}}0.779 +-  0.003\end{tabular}          & \begin{tabular}[c]{@{}l@{}}0.785 +-  0.009\end{tabular}          & \begin{tabular}[c]{@{}l@{}}0.579 +-  0.003\end{tabular}          & \begin{tabular}[c]{@{}l@{}}2.004 +-  0.004\end{tabular}          & \begin{tabular}[c]{@{}l@{}}0.238 +-  0.002\end{tabular}          & \begin{tabular}[c]{@{}l@{}}0.833 +-  0.002\end{tabular}          \\ 
                                                                            & \begin{tabular}[c]{@{}l@{}}\algname -S (0.3)\end{tabular} & \textbf{\begin{tabular}[c]{@{}l@{}}0.785 +-  0.002\end{tabular}} & \textbf{\begin{tabular}[c]{@{}l@{}}0.786 +-  0.004\end{tabular}} & \textbf{\begin{tabular}[c]{@{}l@{}}0.588 +-  0.003\end{tabular}} & \textbf{\begin{tabular}[c]{@{}l@{}}2.048 +-  0.000\end{tabular}} & \textbf{\begin{tabular}[c]{@{}l@{}}0.081 +-  0.001\end{tabular}} & \textbf{\begin{tabular}[c]{@{}l@{}}0.943 +-  0.001\end{tabular}} \\ \midrule
\multirow{3}{*}{NYT}                                                        & BCE                                           & \begin{tabular}[c]{@{}l@{}}0.627 +-  0.001\end{tabular}          & \begin{tabular}[c]{@{}l@{}}0.515 +-  0.002\end{tabular}          & \begin{tabular}[c]{@{}l@{}}0.382 +-  0.001\end{tabular}          & \begin{tabular}[c]{@{}l@{}}2.636 +-  0.001\end{tabular}          & \begin{tabular}[c]{@{}l@{}}0.466 +-  0.001\end{tabular}          & \begin{tabular}[c]{@{}l@{}}0.792 +-  0.000\end{tabular}          \\  
                                                                            & \begin{tabular}[c]{@{}l@{}}\algname -H (0.3)\end{tabular} & \textbf{\begin{tabular}[c]{@{}l@{}}0.656 +-  0.002\end{tabular}} & \begin{tabular}[c]{@{}l@{}}0.591 +-  0.009\end{tabular}          & \textbf{\begin{tabular}[c]{@{}l@{}}0.473 +-  0.006\end{tabular}} & \begin{tabular}[c]{@{}l@{}}2.662 +-  0.002\end{tabular}          & \begin{tabular}[c]{@{}l@{}}0.428 +-  0.004\end{tabular}          & \begin{tabular}[c]{@{}l@{}}0.803 +-  0.001\end{tabular}          \\  
                                                                            & \begin{tabular}[c]{@{}l@{}}\algname -S (0.7)\end{tabular} & \begin{tabular}[c]{@{}l@{}}0.648 +-  0.002\end{tabular}          & \textbf{\begin{tabular}[c]{@{}l@{}}0.607 +-  0.004\end{tabular}} & \begin{tabular}[c]{@{}l@{}}0.468 +-  0.005\end{tabular}          & \textbf{\begin{tabular}[c]{@{}l@{}}2.715 +-  0.005\end{tabular}} & \textbf{\begin{tabular}[c]{@{}l@{}}0.266 +-  0.015\end{tabular}} & \textbf{\begin{tabular}[c]{@{}l@{}}0.886 +-  0.005\end{tabular}} \\ \midrule
\multirow{3}{*}{RCV1}                                                       & BCE                                           & \begin{tabular}[c]{@{}l@{}}0.659 +-  0.006\end{tabular}          & \textbf{\begin{tabular}[c]{@{}l@{}}0.458 +-  0.013\end{tabular}} & \begin{tabular}[c]{@{}l@{}}0.508 +-  0.005\end{tabular}          & \begin{tabular}[c]{@{}l@{}}1.621 +-  0.003\end{tabular}          & \begin{tabular}[c]{@{}l@{}}0.281 +-  0.007\end{tabular}          & \begin{tabular}[c]{@{}l@{}}0.844 +-  0.005\end{tabular}          \\ 
                                                                            & \begin{tabular}[c]{@{}l@{}}\algname -H (0.001)\end{tabular} & \textbf{\begin{tabular}[c]{@{}l@{}}0.675 +-  0.002\end{tabular}} & \begin{tabular}[c]{@{}l@{}}0.444 +-  0.003\end{tabular}          & \textbf{\begin{tabular}[c]{@{}l@{}}0.517 +- 0.001\end{tabular}} & \begin{tabular}[c]{@{}l@{}}1.633 +-  0.002\end{tabular}          & \begin{tabular}[c]{@{}l@{}}0.270 +-  0.003\end{tabular}          & \begin{tabular}[c]{@{}l@{}}0.849 +-  0.001\end{tabular}          \\ 
                                                                            & \begin{tabular}[c]{@{}l@{}}\algname -S (0.1)\end{tabular} & \begin{tabular}[c]{@{}l@{}}0.662 +-  0.002\end{tabular}          & \begin{tabular}[c]{@{}l@{}}0.445 +-  0.008\end{tabular}          & \begin{tabular}[c]{@{}l@{}}0.504 +-  0.004\end{tabular}          & \textbf{\begin{tabular}[c]{@{}l@{}}1.645 +-  0.004\end{tabular}} & \textbf{\begin{tabular}[c]{@{}l@{}}0.222 +-  0.005\end{tabular}} & \textbf{\begin{tabular}[c]{@{}l@{}}0.878 +-  0.003\end{tabular}} \\ \midrule
\multirow{3}{*}{\begin{tabular}[c]{@{}l@{}}FSDK \\ Audio\end{tabular}}      & BCE                                           & \begin{tabular}[c]{@{}l@{}}0.467 +-  0.003\end{tabular}          & \begin{tabular}[c]{@{}l@{}}0.438 +-  0.004\end{tabular}          & \begin{tabular}[c]{@{}l@{}}0.437 +-  0.002\end{tabular}          & \begin{tabular}[c]{@{}l@{}}1.649 +-  0.010\end{tabular}          & \begin{tabular}[c]{@{}l@{}}0.849 +-  0.006\end{tabular}          & \begin{tabular}[c]{@{}l@{}}0.643 +-  0.003\end{tabular}          \\ 
                                                                            & \begin{tabular}[c]{@{}l@{}}\algname -H (0.2)\end{tabular} & \textbf{\begin{tabular}[c]{@{}l@{}}0.471 +-  0.005\end{tabular}} & \textbf{\begin{tabular}[c]{@{}l@{}}0.448 +-  0.002\end{tabular}} & \textbf{\begin{tabular}[c]{@{}l@{}}0.442 +-  0.003\end{tabular}} & \begin{tabular}[c]{@{}l@{}}1.667 +-  0.014\end{tabular}          & \begin{tabular}[c]{@{}l@{}}0.792 +-  0.008\end{tabular}          & \begin{tabular}[c]{@{}l@{}}0.661 +-  0.003\end{tabular}          \\  
                                                                            & \begin{tabular}[c]{@{}l@{}}\algname -S (0.8)\end{tabular} & \begin{tabular}[c]{@{}l@{}}0.445 +-  0.010\end{tabular}          & \begin{tabular}[c]{@{}l@{}}0.431 +-  0.007\end{tabular}          & \begin{tabular}[c]{@{}l@{}}0.421 +-  0.008\end{tabular}          & \textbf{\begin{tabular}[c]{@{}l@{}}1.679 +- 0.024\end{tabular}} & \textbf{\begin{tabular}[c]{@{}l@{}}0.588 +-  0.034\end{tabular}} & \textbf{\begin{tabular}[c]{@{}l@{}}0.744 +- 0.014\end{tabular}} \\ 
\bottomrule
\end{tabular}
}
\caption{This table compares standard binary cross-entropy (BCE) loss for multi-label classification and \algname with $\beta$ values in brackets. The average value over three runs is reported in the table. Most of the experiments have a standard deviation value of <0.005. \algnamehard and \algnamesoft use the severity scores~\eqref{eqn:loss-hard} and~\eqref{eqn:loss-soft} respectively. For each metric \red{$\uparrow$} (resp. \blue{$\downarrow$}) indicates that higher (resp. lower) is better. As we can see from the results, both \algnamehard and \algnamesoft substantially outperform BCE on hierarchy-specific metrics such as LCA, AHC, and HP and standard metrics such as AUPRC P@5 and F1@5, demonstrating the utility of hierarchy information.}
\label{main_table}
\end{table}
Table~\ref{main_table} compares the performance of \algnamehard and \algnamesoft, which use severity metrics~\eqref{eqn:loss-hard} and~\eqref{eqn:loss-soft} respectively with the standard BCE loss approach. The results clearly demonstrate the superior performance of both \algnamehard and \algnamesoft compared to BCE.
We see substantial improvements in standard metrics such as AUPRC, P@5, and F1@5 and hierarchical metrics such as LCA, AHC, and HP on all of the datasets. We see the most significant improvements on the OpenImages dataset, where \algnamehard achieves a $64.91\%$ relative improvement over BCE in AUPRC. While there is no clear trend between \algnamehard and \algnamesoft on the standard metrics (i.e., AUPRC, P@5, and F1@5), \algnamesoft consistently outperforms \algnamehard on the hierarchical metrics (i.e., LCA, AHC, and HP). In the following experiments, for any given dataset, we consider the version of \algname that gives better AUPRC. So, we use \algnamesoft for Food 201, COCO and \algnamehard for the others.

\begin{table}[t]
\resizebox{\linewidth}{!}{
\begin{tabular}{llllllll}
\toprule
\textbf{Label distribution} & \textbf{Experiment\textbackslash{}Dataset} & \multicolumn{1}{l}{\textbf{OpenImages}} & \multicolumn{1}{l}{\textbf{COCO}} & \multicolumn{1}{l}{\textbf{Food201}} & \multicolumn{1}{l}{\textbf{RCV1}} & \multicolumn{1}{l}{\textbf{NYT}} & \multicolumn{1}{l}{\textbf{FSDK}} \\ \midrule
Top 20\%       & BCE                              & 0.537                                     & 0.678                              & 0.574                                 & 0.897                              & 0.886                             & 0.435                              \\ \midrule
Top 20\%         & CHAMP                          & 0.584                                    & 0.682                              & 0.593                                 & 0.901                              & 0.612                             & 0.442                              \\ \midrule
Bottom 20\%       & BCE                               & 0.146                                    & 0.639                              & 0.121                                 & 0.337                              & 0.203                             & 0.520                              \\ \midrule
Bottom 20\%      & CHAMP                             & 0.612                                    & 0.675                              & 0.138                                 & 0.357                              & 0.459                             & 0.573                              \\ 
\bottomrule
\end{tabular}
}
\caption{AUPRC computed on the top $20\%$ and bottom $20\%$ labels, sorted according to the number of examples per label, for both BCE and \algname on the six datasets. We see that \algname achieves much more improvement on the bottom $20\%$ labels than on the top $20\%$ labels.}
\label{tab:skew}
\end{table}
\textbf{Why does \algname improve?}: Intuitively, additional side information, such as hierarchy, is likely to help those labels which have very few examples. To verify this intuition, we order the labels in decreasing order of the number of examples and consider the top $20\%$ labels with the most number of examples and the bottom $20\%$ labels with the least number of examples.
Table~\ref{tab:skew} presents the AUPRC numbers for these subsets of labels, achieved by both BCE and \algname on all the six datasets. The results confirm this intuition, as we can see that the AUPRC improvements achieved by \algname for the bottom $20\%$ labels are substantially larger than that for the top $20\%$.

\subsection{Comparison to single-label hierarchical training}
\label{sec:results-singlelabel}


This section compares single-label hierarchical training with our HMC framework to highlight the importance of multilabel training. For each sample $x_i$ with label set $S_i$, we construct a new dataset such that sample $x_i$ is repeated in the dataset $\abs{S_i}$ times each with a different label. Thus, we convert the multilabel problem into a single label problem where sample-label pairs include $(x_i,j)$ where $j \in S_i$. We denote this experiment as M2S. As we can see in Table \ref{m2stable}, single label hierarchical training using the same distance-based loss formulation as CHAMP leads to lower AUPRC performance. We hypothesize that since \algname is trained with implicit multilabel relationships, it can achieve better AUPRC scores, as seen in Table~\ref{m2stable}. Both M2S and \algname have similar performance on hierarchical metrics since they train with the same hierarchy information and loss function. Our comparison of \algname with BCE and M2S helps identify the contribution of individual gains from hierarchy information and co-occurrence data for each dataset.


\begin{table}
\centering
\resizebox{\linewidth}{!}{
\begin{tabular}{lllll}
\toprule
\textbf{Experiment\textbackslash{}Dataset} & \textbf{Food201}        & \textbf{COCO}           & \textbf{RCV1}           & \textbf{FSDK}           \\ 
\midrule
M2S                               & 0.573 +- 0.004         & 0.744 +- 0.002         & 0.674 +- 0.003        & 0.449 +- 0.009       \\ 
CHAMP                             & \textbf{0.593 +- 0.003} & \textbf{0.785 +- 0.002 } & \textbf{0.675 +- 0.002} & \textbf{0.471 +- 0.005 } \\ 
\bottomrule
\end{tabular}
}
\caption{Effect of multilabel information compared to single label information on AUPRC. M2S replaces the multilabel dataset with single label data points and uses BCE loss and \algname formulation. \algname, on the other hand, works directly with the multilabel data. We see that \algname consistently outperforms M2S.}
\label{m2stable}
\end{table}

\subsection{Robustness to adding noise}
\begin{figure*}[t]
	\centering
	\includegraphics[width=1\textwidth]{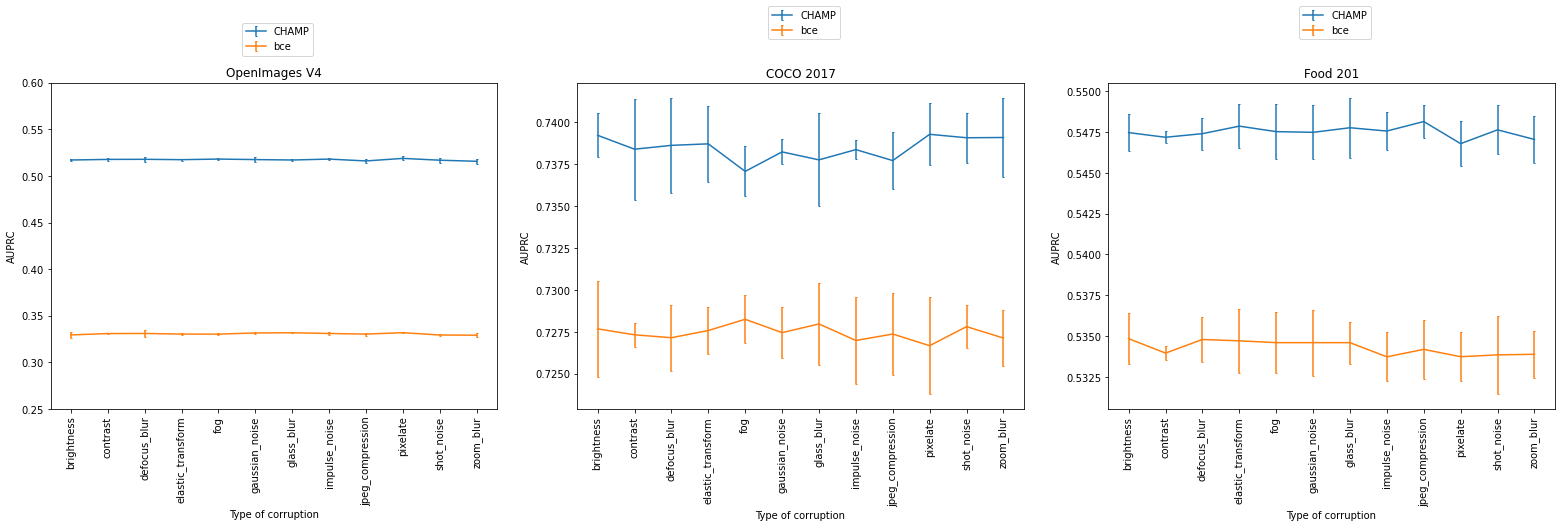}
	\caption{Effect of \algname on robustness to twelve popular corruptions to images. We see that on all three datasets, \algname obtains improved robustness compared to BCE, owing to the semantic/hierarchical information imparted to it during training.}
	\label{fig:Robustness}
\end{figure*}
In this section, we show the effect of \algname on the robustness of the learned models to twelve popular corruptions for visual data: Gaussian noise, shot noise, impulse noise, defocus blur, glass blur, zoom blur, fog, brightness, contrast, elastic transform, pixelate and jpeg compression~\cite{https://doi.org/10.48550/arxiv.1807.01697}. We train on the original image training sets (OpenImages, Food 201, and COCO) but add the corruptions to the test sets with five levels of severity. Figure~\ref{fig:Robustness} presents AUPRC on the test set averaged over five levels of severity of noise. The results demonstrate that \algname produces more robust models owing to the additional semantic/hierarchical information imparted during training.

\subsection{Training with less data}
\begin{figure*}[ht]
	\centering
	\includegraphics[width=1\textwidth, height=30mm]{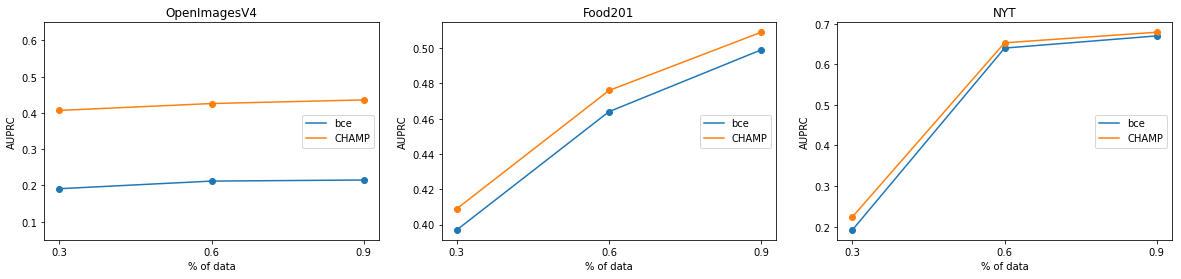}
	\caption{AUPRC achieved by BCE and \algname for varying fractions of training data. We see that \algname gives consistent improvements over BCE at various fractions of training data.}
	\label{fig:percent_data}
\end{figure*}
In this section, we investigate the effectiveness of \algname at various levels of training sample size. Figure~\ref{fig:percent_data} presents the AUPRC achieved by both BCE as well as \algname when different fractions of training data are used across the three image datasets. The results demonstrate that \algname substantially improves a broad range of training data sizes.

\subsection{Augmenting \algname with other multi-label loss functions}
\label{sec:results-augmentotherloss}
This section demonstrates how \algname can augment state-of-the-art multilabel loss functions. In particular, we consider the asymmetric loss function introduced in~\cite{https://doi.org/10.48550/arxiv.2009.14119} and augment it with \algname to obtain \algnameasymm (see Supplementary material for the precise formulation). Table~\ref{tab:asymm} presents these results. The results demonstrate that while \algname suffers a slight drop in AUPRC, it obtains significant improvements on hierarchical metrics such as LCA. The results support our claim to provide a generic framework for existing and future multilabel classification methods to be augmented with inter-label relationships. We also look forward to extensions of our work to augment single-label hierarchical methods. 
\begin{table}
\centering
\resizebox{\linewidth}{!}{
\begin{tabular}{llllllll}
\toprule
\textbf{Metric}        & \textbf{Experiment\textbackslash{}Dataset} & \textbf{OpenImages} & \textbf{COCO}  & \textbf{Food201} & \textbf{NYT}   & \textbf{RCV1}  & \textbf{FSDK}  \\ \midrule
\multirow{2}{*}{AUPRC} & Asymmetric                        & \textbf{0.435 +- 0.005 }      & 0.751 +- 0.010          & \textbf{0.590 +- 0.001 }   & \textbf{0.655 +- 0.001 } & \textbf{0.676 +- 0.001 } & \textbf{0.461 +- 0.005 } \\ 
                       & \algnameasymm                           & 0.407 +- 0.003               & \textbf{0.768 +- 0.002 } & 0.584 +- 0.001             & 0.643 +- 0.001           & 0.662 +- 0.001           & 0.458 +- 0.007          \\ \midrule
\multirow{2}{*}{LCA}   & Asymmetric                                 & 0.777 +- 0.006               & 1.480 +- 0.001          & 0.824 +- 0.018            & 2.324 +- 0.013           & 1.426 +- 0.004          & 1.322 +- 0.019          \\ 
                       & \algnameasymm                           & \textbf{0.887 +- 0.006 }      & \textbf{1.924 +- 0.006 } & \textbf{1.114 +- 0.009 }   & \textbf{2.599 +- 0.005 } & \textbf{1.470 +- 0.005 } & \textbf{1.511 +- 0.006 } \\ 
\bottomrule                       
\end{tabular}
}
\caption{AUPRC and LCA metrics achieved by the asymmetric multilabel loss~\cite{https://doi.org/10.48550/arxiv.2009.14119} as well as \algnameasymm. We can see that while the asymmetric loss achieves slightly better AUPRC, \algnameasymm achieves better hierarchical metrics.}
\label{tab:asymm}
\end{table}

\section{Related Work}
\label{sec:related}

Previous work in multi-label classification has been around exploiting label correlation via graph neural networks \cite{https://doi.org/10.48550/arxiv.1904.03582}, \cite{https://doi.org/10.48550/arxiv.1902.09720}, \cite{8784938} or word embeddings \cite{https://doi.org/10.48550/arxiv.1904.03582}, \cite{https://doi.org/10.48550/arxiv.1911.09243}. Others are based on modeling image parts and attentional
regions \cite{https://doi.org/10.48550/arxiv.1912.07872}, \cite{Gao_2021}, \cite{https://doi.org/10.48550/arxiv.1711.02816}, \cite{https://doi.org/10.48550/arxiv.2012.02994} as well as using recurrent neural networks \cite{NIPS2017_2eb5657d}, \cite{https://doi.org/10.48550/arxiv.1604.04573}, embedding space constraints, \cite{https://doi.org/10.48550/arxiv.2106.11596}, region sampling \cite{https://doi.org/10.48550/arxiv.1702.05891}, \cite{Gao_2021} and cross-attention \cite{https://doi.org/10.48550/arxiv.2111.12933}. There has also been recent interest in multi-label text classification \cite{10.1145/3077136.3080834}, \cite{Pal_2020}, \cite{6121678}, \cite{https://doi.org/10.48550/arxiv.1812.09910}, \cite{https://doi.org/10.48550/arxiv.1909.04176}

Recently, progress has also been made in incorporating hierarchical knowledge to single label classifiers to add additional semantics to the models' learning capabilities such that even when the model makes mistakes dues to data ambiguities, it is able to make semantically better mistakes. Hierarchical information is important in many other applications such as food recognition \cite{https://doi.org/10.48550/arxiv.2012.03368}, \cite{10.1145/2964284.2967205}, protein function prediction \cite{pmlr-v8-cesa-bianchi10a}, \cite{10.5555/3104482.3104485}, \cite{TRIGUERO2016170}, \cite{10.1007/978-3-642-40991-2_25}, \cite{Otero2010AHM}, \cite{doi:10.1080/13102818.2018.1521302}
, image annotation \cite{DIMITROVSKI20112436}, text classification \cite{Mao_2019}, \cite{10.5555/1248547.1248606}, \cite{shen-etal-2021-taxoclass}. Some major approaches include imposing logical constraints \cite{giunchiglia2020coherent}, using hyperbolic embeddings  \cite{dhall2020hierarchical}, prototype learning \cite{garnot2021leveraging}, label smearing and soft labels, loss modifications \cite{https://doi.org/10.48550/arxiv.1912.09393}, multiple learning heads for different levels of the hierarchy \cite{app11083722}, hierarchical post-processing \cite{https://doi.org/10.48550/arxiv.2104.00795} and others \cite{10.1007/978-3-030-31654-9_5}, \cite{NIPS2011_5a4b25aa}, \cite{pmlr-v28-agarwal13}.

While there has been some previous literature in HMC in the domain of protein function prediction and online advertising systems \cite{agrawal2013multi-label}, there is limited work in the deep learning era on modalities like audio and images. There has also been some recent work in hierarchical multilabel text classification \cite{https://doi.org/10.48550/arxiv.2101.04997}, \cite{Mao_2019}, \cite{https://doi.org/10.48550/arxiv.1905.10802}, \cite{shen-etal-2021-taxoclass},  \cite{DAISEY2020104177}. Local classification approaches \cite{CesaBianchi2006HierarchicalCC} train a set of classifiers at each level
of the hierarchy. However, it has also been argued \cite{6121678} that it is impractical to
train separate classifiers at each level due to the several assumptions involved in semantics across siblings or nodes at the same level. Local Classifiers can be further divided: Local classifier per level (LCL) \cite{6121678}, Local classifier per node(LCN) \cite{tprhe} and local classifier per parent node (LCP). On the other
hand, Global approaches \cite{5360345} train a single classifier
that factors in the complete tree. Unlike
the local approach, global approaches do not suffer from error propagation. Local methods are better at capturing
label correlations, whereas global methods are less computationally expensive. Researchers recently tried to use a
hybrid loss function associated with specifically designed
neural networks \cite{10.1145/3178876.3186005}. The archetype of HMCN-F \cite{pmlr-v80-wehrmann18a}
employs a cascade of neural networks, where each neural
network layer corresponds to one level of the label hierarchy. Such neural network architectures generally require
all the paths in the label hierarchy to have the same length,
which limits their application. \cite{https://doi.org/10.48550/arxiv.2010.10151} approach HMC by using parent and child probabilities constraints. However, existing work is majorly focused on just using hierarchy to improve learning but has still not focused on the notion of better mistakes in the multi-label domain, motivating us to work on CHAMP.

\section{Learnings and Conclusion}
\label{sec:disc}
\textbf{Summary}: In this work, we consider the problem of hierarchical multilabel classification. While there is a rich literature on both multilabel and hierarchical classification individually, there are few works that develop a general method for hierarchical multilabel classification. In this paper, we develop one such algorithm -- \algname -- and, through experiments on six diverse datasets across vision, text, and audio, demonstrate that \algname provides improvements not only on hierarchical metrics but also on standard metrics like AUPRC

\textbf{Discussion}: Supervised deep neural networks rely on large amounts of data to generalize well to unseen data. While a typical way to increase the data is to collect more examples with supervised labels, there is also a growing appreciation for capturing labels with richer semantics and richer annotations. Hierarchical and multiple label annotations can cover many relations like part-of, is-a, and similar-to. We encourage the community to invest more in better labeling procedures by showcasing downstream gains such as training with less data, making better mistakes, and robustness to noise. Though we envisage a natural extension of our algorithm to DAG (directed acyclic graphs), it would need more work to make semantic aware mistakes in datasets with generic graphical label relationships. Our methods also set the platform and extend to more complex learning algorithms like detection \cite{https://doi.org/10.48550/arxiv.2107.13627}, segmentation, image retrieval, and ranking to make better mistakes. Even contrastive learning approaches like triplet loss and object detection \cite{https://doi.org/10.48550/arxiv.2107.13627} can naturally extend to our approach and benefit from hierarchy and co-occurrence information to adjust the loss functions to make better mistakes. Hierarchy relations are more relevant when the feature representation similarity does not align with semantic similarity, and the cost of making mistakes is high. For example, a black slate and a tv monitor look visually similar but are semantically different. We see $2.74\%$ and $0.9\%$ percentage drop in AUPRC when trained with random tree structures on OpenImages and COCO datasets. This occurs due to conflict in semantic and visual similarities.
Similarly, it is crucial to evaluate the benefits \& cost of adding multilabel information. Our work encourages the community to invest in richer annotations with the promise of higher performance and other advantages. We conclude our CHAMP work by setting the first platform in the journey to learn making better mistakes in the context of rich label relationships.

\newpage
\bibliographystyle{unsrt}
\bibliography{main}

\clearpage 

\appendix
\section{Appendix}

\begin{figure*}[ht]
	\centering
	\includegraphics[width=1\textwidth, height=150mm]{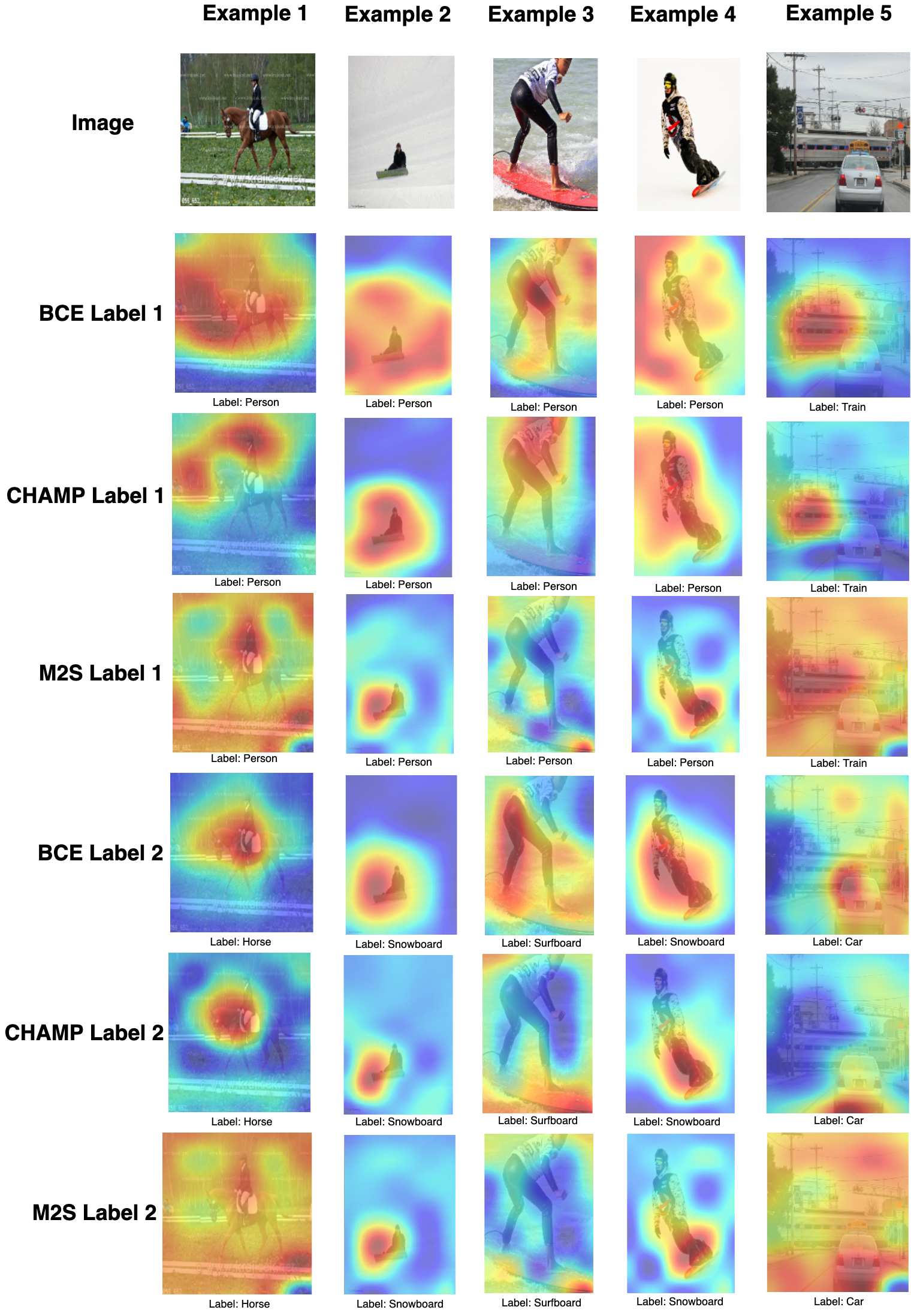}
	\caption{Comparing CHAMP vs. BCE and M2S across highly co-occurring classes to demonstrate the overreliance of baseline multilabel classification models on co-occurrence information}
	\label{fig:matrix}
\end{figure*}

\subsection{Understanding CHAMP better}
 Our experiments show that CHAMP leads to more robust models. We design further experiments to understand the reasons that make CHAMP learn richer and more semantic aware representations. The following section explores how hierarchy complements multilabel information and vice-versa. We explore the benefits of CHAMP compared with baseline multilabel models among image datasets. Further research is needed to understand similar effects of HMC modeling on text datasets in the future. 

\subsubsection{Reliance on co-occurrence}
Multi-label classification implicitly uses co-occurrence of labels to make model predictions. However, this over-reliance on co-occurrence information can be detrimental as it can lead to the model learning spurious correlations for certain labels. We demonstrate this by plotting the class activation maps (Grad-CAM \cite{Selvaraju_2019}) for each model for a few examples from the image datasets used in our work in Figure \ref{fig:matrix}. In example 1, the co-occurrence probability of class 'person' in the presence of the class 'horse' is 69.53\%. Similarly, the co-occurrence probability of class 'person' in the presence of class 'surfboard' and 'snowboard' is 95.3\% and 97.95\% respectively, and 'traffic light' in the presence of 'car' is 61.25\%. We see from figure \ref{fig:matrix} that the BCE model is over-reliant on this information as the activation maps for BCE focus on 'person' extensively too while predicting horse, snowboard, and surfboard, respectively. CHAMP learns better features forced with hierarchical loss to learn features common to the parent.
Similarly, the baseline BCE model attends to both car and traffic lights while predicting car. However, CHAMP can use hierarchical relationships to learn richer representations that are not over-reliant on the co-occurrence of labels. We intend to conduct further research to strengthen our hypothesis that HMC modeling leads to robust feature representation in future work. In the initial explorations, we provide insights into CHAMP's better performance in scenarios like less training data and adding noise to images.

\begin{figure*}[ht]
	\centering
	\includegraphics[width=1\textwidth, height=50mm]{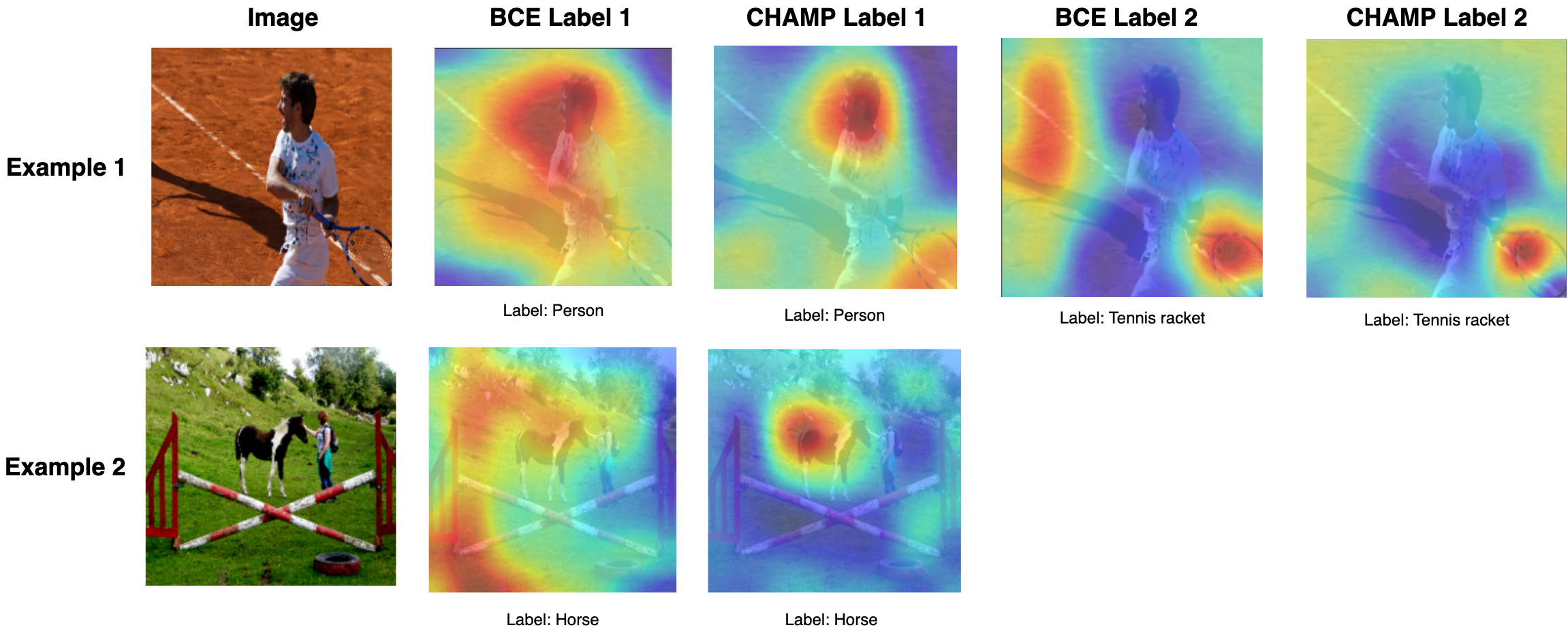}
	\caption{Comparing CHAMP vs. BCE about reliance on background information}
	\label{fig:bg}
\end{figure*}

\subsubsection{Reliance on background}
It has also been studied that deep neural networks are overly reliant on background \cite{https://doi.org/10.48550/arxiv.2006.09994}. Hierarchy helps dilate this over-reliance by providing further semantics about the image and labels. Since a sample is not just defined by leaf nodes but instead the complete ancestral path, CHAMP can differentiate between the foreground and background better with the help of hierarchical information. We demonstrate this observation in Figure \ref{fig:bg}. The baseline BCE model extensively attends to the background while predicting 'tennis racket' and 'horse' in examples 1 and 2. However, CHAMP can learn richer representations that focus majorly on the actual object.

\begin{figure*}[ht]
	\centering
	\includegraphics[width=1\textwidth, height=25mm]{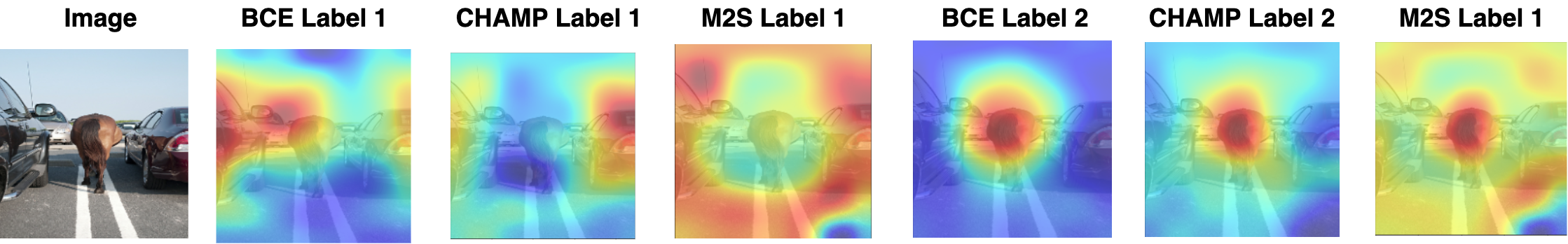}
	\caption{Comparing CHAMP vs. BCE and M2S experiments across OOD / low co-occurrence labels.}
	\label{fig:ood}
\end{figure*}

\subsection{Hierarchy helps in low co-occurrence cases}
Out of distribution examples and adversarial attacks can exploit the over-reliance on high co-occurring labels. In this section, we take examples from low co-occurring labels and show that CHAMP focuses on the correct regions compared to the baseline. As can be seen in Figure \ref{fig:ood}, we take the example of a cow and car, two classes with a very low co-occurrence percentage in the dataset (4.59\%). Hierarchy can add value to multi-label learning since it provides another dimension (semantic relationships) other than co-occurrence relationships. The CHAMP model demonstrates better feature representation for car and cow owing to hierarchical information while training.  

\subsection{Importance of training with multiple labels simultaneously}
We can see in Figure \ref{fig:matrix} that the activation maps for the M2S experiments are unable to attend to individual objects in the image distinctly. This may be because the model is trained with different labels individually in each iteration, thereby preventing it from learning with the confidence that multiple labels could be present. CHAMP brings better attention by adding multi-label loss for the model to attribute the presence of multiple objects in an image. 

\subsection{Localising improvements}
This section dissects where we find improvements with CHAMP compared to the baseline BCE model. We demonstrate this using the COCO dataset for brevity, although similar trends are observed for other datasets. In Table~\ref{delta_level}, we observe that the most significant mean improvements are observed at the deepest level in the tree (leaf nodes) with an increasing trend from root to leaf nodes. Figure \ref{fig:rank} shows that CHAMP improves performance for low ranked nodes more, that is, labels with less number of samples. This is intuitive as hierarchical and multilabel information together help in improving the low data problem, as we have observed in our experiments with less data.

\begin{table}[ht]
\centering
\resizebox{\linewidth}{!}{
\begin{tabular}{rrrr}
\toprule
\multicolumn{1}{l}{\textbf{Level}} & \multicolumn{1}{l}{\textbf{Mean \% AUPRC difference (CHAMP, BCE)}} & \multicolumn{1}{l}{\textbf{Mean \% data per node}} & \multicolumn{1}{l}{\textbf{Number of nodes}} \\ \midrule
1                                    & -0.01\%                                                             & 9.75                                        & 2                                             \\ 
2                                    & 0.79\%                                                              & 2.96                                        & 12                                            \\ 
3                                    & 2.41\%                                                              & 0.56                                        & 80                                            \\
\bottomrule
\end{tabular}
}
\caption{\label{delta_level}Comparing the mean \% AUPRC improvement of CHAMP vs BCE across different levels of the tree)}
\end{table}

\begin{figure*}[ht]
	\centering
	\includegraphics[width=1\textwidth, height=80mm]{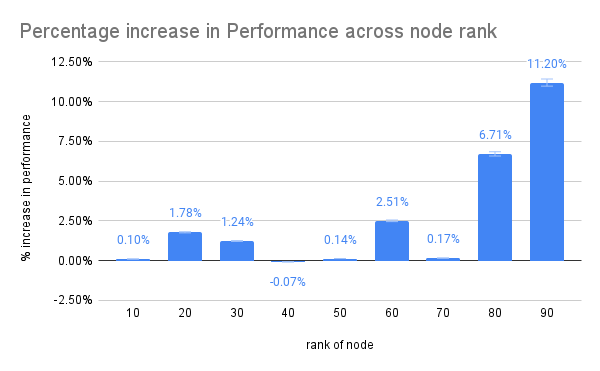}
	\caption{Comparing the \% increase in AUPRC vs. rank of the node with respect to the number of samples per label}
	\label{fig:rank}
\end{figure*}

\subsection{Impact of scaling}
In this section, we study the impact of different scales by which we can increase the penalty for the severity of the mistakes. Table \ref{scaling_table_complete} shows that while hierarchical metrics can further increase by scaling the false-negative term more, AUPRC scores can take a hit. 

\begin{table}
\centering
\begin{tabular}{llllllll}
\toprule
\textbf{Dataset}                                                                     & \textbf{Experiment}                                               & \textbf{AUPRC} & \textbf{P@5}   & \textbf{F1@5}  & \textbf{LCA}   & \textbf{AHC}    & \textbf{HP}    \\ \midrule
\multirow{3}{*}{\begin{tabular}[c]{@{}l@{}}Open Images v4\end{tabular}} & \begin{tabular}[c]{@{}l@{}}beta*d\end{tabular} & \textbf{0.559} & \textbf{0.283} & 0.375          & 1.920          & 0.454           & 0.726          \\
                                                                                     & (1+beta)*d                                               & 0.554          & 0.275          & 0.366          & \textbf{1.964} & \textbf{0.393}  & \textbf{0.757} \\ 
                                                                                     & beta*d*d                                                 & 0.559          & 0.283          & \textbf{0.376} & 1.825          & 0.488           & 0.701          \\ \midrule
\multirow{3}{*}{\begin{tabular}[c]{@{}l@{}}Food 201\end{tabular}}         & \begin{tabular}[c]{@{}l@{}}beta*d\end{tabular} & \textbf{0.593} & 0.528          & 0.486          & 1.557          & 0.316           & 0.861          \\ 
                                                                                     & (1+beta)*d                                               & 0.591          & 0.527          & 0.482          & \textbf{1.599} & \textbf{0.2433} & \textbf{0.891} \\ 
                                                                                     & beta*d*d                                                 & 0.589          & \textbf{0.526} & \textbf{0.478} & 1.558          & 0.3075          & 0.863          \\ \midrule
\multirow{3}{*}{COCO}                                                       & \begin{tabular}[c]{@{}l@{}}beta*d\end{tabular} & 0.785          & 0.786          & 0.588          & 2.048          & 0.081           & 0.943          \\ 
                                                                                     & (1+beta)*d                                               & \textbf{0.787} & \textbf{0.788} & \textbf{0.589} & \textbf{2.054} & \textbf{0.058}  & \textbf{0.958} \\ 
                                                                                     & beta*d*d                                                 & 0.783          & 0.783          & 0.5825         & 2.046          & 0.075           & 0.945          \\ \midrule
\multirow{3}{*}{NYT}                                                        & \begin{tabular}[c]{@{}l@{}}beta*d\end{tabular} & \textbf{0.648} & 0.607          & \textbf{0.468} & 2.715          & 0.266           & 0.886          \\ 
                                                                                     & (1+beta)*d                                               & 0.644          & \textbf{0.608} & 0.464          & \textbf{2.721} & \textbf{0.248}  & \textbf{0.894} \\ 
                                                                                     & beta*d*d                                                 & 0.644          & 0.597          & 0.467          & 2.707          & 0.276           & 0.881          \\ \midrule
\multirow{3}{*}{RCV1}                                                       & \begin{tabular}[c]{@{}l@{}}beta*d\end{tabular} & \textbf{0.675} & 0.444          & \textbf{0.517} & 1.633          & 0.270           & 0.849          \\ 
                                                                                     & (1+beta)*d                                               & 0.646          & 0.437          & 0.490          & \textbf{1.637} & \textbf{0.242}  & \textbf{0.864} \\ 
                                                                                     & beta*d*d                                                 & 0.657          & \textbf{0.445} & 0.500          & 1.627          & 0.273           & 0.845          \\ \midrule
\multirow{3}{*}{\begin{tabular}[c]{@{}l@{}}FSDK Audio\end{tabular}}      & \begin{tabular}[c]{@{}l@{}}beta*d\end{tabular} & \textbf{0.471} & 0.448          & \textbf{0.442} & \textbf{1.667} & 0.792           & 0.661          \\ 
                                                                                     & (1+beta)*d                                               & 0.462          & 0.431          & 0.433          & 1.674          & \textbf{0.761}  & \textbf{0.682} \\  
                                                                                     & beta*d*d                                                 & 0.466          & \textbf{0.448} & 0.437          & 1.661          & 0.784           & 0.660          \\ 
\bottomrule
\end{tabular}
\caption{\label{scaling_table_complete} Comparing different scaling strategies and the trade-off between multilabel and hierarchical metrics}
\end{table}


\subsection{Details of experiments for reproducibility}
We use the standard train and test splits for all the six datasets given by their authors. Food201, MS-COCO, and FSDKaggle2019 datasets only provide leaf-level annotations. We thus assume that if a given sample is labeled with a child node, it should also be labeled with its parent ancestry. Thus, we augment the labels for Food201, MS-COCO, and FSDKaggle2019 datasets with parent labels. 

For the New York Times dataset, the label 'Columns' appears at different levels of the hierarchy, making it a Directed Acyclic Graph (DAG). Thus, we do not consider this label as in \cite{https://doi.org/10.48550/arxiv.2101.04997}. Moreover, for the Reuters corpus volume 1 dataset, two classes in the test set do not have any samples in the training set. We do not modify this scenario and use the same train/test split given by the dataset authors. For the NYT dataset, we use the entire training data as compared to the sampled 25,279 samples and 'full\_text' data as compared to just the 'lead\_paragraph' used in \cite{Mao_2019}.

\newpage
\subsection{Code}
Below we provide the code for CHAMP-hard. The extension to CHAMP soft is trivial by removing the hard assignment step and adding appropriate weights and will be made publicly available soon.
\begin{lstlisting}
def champ_hard_loss(dict_idx2labels, dict_kid2parent, n_labels, beta):
  """Implementation of CHAMP hard loss.

  Args:
    dict_idx2labels: dict mapping tree indices (int) to labels (string)
    dict_kid2parent: Dictionary representing tree in kid -> parent format.
    n_labels: number of labels
    beta: hyperparam to quantify impact of distance

  Returns:
    hard loss: Computed loss value of champ hard loss.

  """

  normalised_distance = np.zeros([n_labels, n_labels], dtype=np.float32)
  distance_matrix = np.zeros([n_labels, n_labels], dtype=np.float32)
  list_dist = []

  for i in range(n_labels):
    for j in range(n_labels):
      distance_matrix[i][j] = get_distance(dict_idx2labels[i],
                                           dict_idx2labels[j],
                                           dict_kid2parent)
      list_dist.append(distance_matrix[i][j])

  max_dist = max(list_dist)

  for i in range(n_labels):
    for j in range(n_labels):
      normalised_distance[i][j] = -1 + ((distance_matrix[i][j] / (max_dist + 1e-15)) + 1) * ((distance_matrix[i][j] / (max_dist + 1e-15)) + 1)

  distance_matrix = tf.Variable(
      distance_matrix, dtype=tf.float32, trainable=False)
  normalised_distance = tf.Variable(
      normalised_distance, dtype=tf.float32, trainable=False)

  @tf.function
  @tf.__internal__.dispatch.add_dispatch_support
  def hard_loss(y_true, y_pred):
    """Wrapper function to compute champ hard.

    Args:
      y_true: Binarised ground truth vector (Batch x total nodes)
      y_pred: Prediction probabilities (Batch x total nodes)

    Returns:
      Champ hard loss
    """

    ones_tensor = tf.ones([tf.shape(y_true)[0], n_labels], tf.float32)
    distance = tf.expand_dims(distance_matrix, 0)

    # Add by 1 in order to avoid edge cases of minm distance
    # since distance[i][j] = 0
    distance = tf.where(distance > -1, distance + 1, 0)

    # mask distance matrix by ground truth values
    distance = tf.multiply(tf.expand_dims(y_true, 1), distance)

    # masked values in above step will be set to 0. In order to compute minm
    # later, we reset those values to a high number greater than max distance
    distance = tf.where(
        tf.cast(distance, tf.float32) < 1., float(max_dist+2),
        tf.cast(distance, tf.float32))

    # Setting indices with minm values in a column to 1 and others to 0.
    # such that for row i and column j, if distance[i][j] = 1, then pred label i
    # is mapped to ground truth value j
    distance = tf.where(
        distance == tf.math.reduce_min(distance, 2, (distance.shape[0], 1)), 1,
        0)

    # Refill our concerned binarised values (when distance is 1) with their
    # respective normalised distances
    distance = tf.where(distance > 0, (tf.expand_dims(normalised_distance, 0)),
                        0)

    # Modify distance according to how much impact we want from distance penalty
    # in loss calculation
    distance = tf.where(distance != 0., beta*(distance) +1., 0.)

    # Computing (1 - p) [part Misprediction term]
    term1 = tf.expand_dims(tf.subtract(ones_tensor, y_pred), 1)

    # Computing log (1-p) [part of Misprediction term]
    term1 = tf.where(term1 != 0, -tf.math.log(term1 + 1e-15), 0)

    # Computing log (p) [part of correct prediction term]
    term2 = tf.where(y_pred != 0, -tf.math.log(y_pred + 1e-15), 0)#*(alpha)

    # Computing binarised matrix with indices of correct predictions as 1
    correct_ids = tf.multiply(
        tf.expand_dims(y_true, 1), tf.expand_dims(tf.eye(n_labels), 0))

    # Computing loss
    loss = tf.squeeze(tf.matmul(term1, distance)) + tf.squeeze(
        tf.matmul(tf.expand_dims(term2, 1), correct_ids))
    
    return tf.reduce_mean(tf.reduce_sum(loss, 1))

  return hard_loss

\end{lstlisting}

\end{document}